\documentclass[letterpaper, 10 pt, conference]{ieeeconf}  

\IEEEoverridecommandlockouts                              

\usepackage{cite}
\usepackage{amsmath,amssymb,amsfonts}
\usepackage{algorithmic}
\usepackage{graphicx}
\usepackage{textcomp}
\usepackage{xcolor}
\usepackage{multirow}

\usepackage{tikz}
\usepackage{pgfplots}
\usepackage[per-mode=fraction, detect-weight=true]{siunitx}
\usepackage[hidelinks]{hyperref}
\usepackage{subcaption}
\usepackage{multirow, graphicx}

\title{\LARGE \bf
Three-Dimensional Vehicle Dynamics State Estimation for High-Speed Race Cars under varying Signal Quality
}

\author{Sven Goblirsch$^{1}$, Marcel Weinmann$^{1}$ and Johannes Betz$^{2}$ 
\thanks{*This work is a result of the joint research project ATLAS-L4. The project is supported by the German Federal Ministry for Economic Affairs and Climate Action (BMWK), based on a decision of the German Bundestag. The author is solely responsible for the content of this publication.} 
\thanks{$^{1}$S. Goblirsch and M. Weinmann are with the Department of Mobility Systems Engineering, TUM School of Engineering and Design, Technical University of Munich, 85748 Garching, Germany}
\thanks{$^{2}$J. Betz is with the Professorship of Autonomous Vehicle Systems, TUM School of Engineering and Design, Technical University of Munich, 85748 Garching, Germany; Munich Institute of Robotics and Machine Intelligence (MIRMI). \newline{}
Corresponding author: \href{mailto:sven.goblirsch@tum.de}{sven.goblirsch@tum.de}}
}

\begin{document}

\onecolumn

\begin{center}
    \textcopyright \ 2024 IEEE. Personal use of this material is permitted. Permission from IEEE must be obtained for all other uses, including reprinting/republishing this material for advertising or promotional purposes, collecting new collected works for resale or redistribution to servers or lists, or reuse of any copyrighted component of this work in other works.
\end{center}

\twocolumn

\maketitle
\thispagestyle{empty}
\pagestyle{empty}

\begin{abstract}

This work aims to present a three-dimensional vehicle dynamics state estimation under varying signal quality. 
Few researchers have investigated the impact of three-dimensional road geometries on the state estimation and, thus, neglect road inclination and banking.
Especially considering high velocities and accelerations, the literature does not address these effects.
Therefore, we compare two- and three-dimensional state estimation schemes to outline the impact of road geometries.
We use an Extended Kalman Filter with a point-mass motion model and extend it by an additional formulation of reference angles.
Furthermore, virtual velocity measurements significantly improve the estimation of road angles and the vehicle's side slip angle.
We highlight the importance of steady estimations for vehicle motion control algorithms and demonstrate the challenges of degraded signal quality and Global Navigation Satellite System dropouts. 
The proposed adaptive covariance facilitates a smooth estimation and enables stable controller behavior.
The developed state estimation has been deployed on a high-speed autonomous race car at various racetracks. Our findings indicate that our approach outperforms state-of-the-art vehicle dynamics state estimators and an industry-grade Inertial Navigation System.
Further studies are needed to investigate the performance under varying track conditions and on other vehicle types.
\end{abstract}

\section{Introduction}
\label{sec:Introduction}

An accurate state estimation is crucial for the safe operation of autonomous vehicles. 
The field of autonomous racing~\cite{betz_survey_2022} inherits even higher demands towards the robustness and accuracy of the estimation due to high velocities combined with narrow tracks.
Further, imprecise localization leads to deviations from the optimal race line and, thus, higher lap times.
Besides, many control approaches incorporate vehicle side slip in the prediction model, requiring precise lateral velocity estimates~\cite{kabzan_learning-based_2019,raji_tricycle_nodate}.
However, external factors such as road inclination, banking, or degraded signal quality complicate an accurate estimation.

In the following, we present and evaluate a three-dimensional state estimation approach considering its longitudinal and lateral speed estimation and localization capabilities.
We compare the results of different algorithm enhancements, including reference angles and various virtual speed measurements, with a state-of-the-art state estimation and an industry-grade inertial navigation system (INS).
Thereby, we consider various challenges faced during the testing sessions and the final races of the INDY Autonomous Challenge~\cite{mitchell2024} at the Autodromo di Monza (MON) and the Las Vegas Motorspeedway (LVMS).
These challenges include banking angles of up to \SI{20}{\degree} at LVMS and degraded Global Navigation Satellite System (GNSS) coverage up to complete dropouts at MON. The presented state estimation improvements have been successfully used in the TUM Autonomous Motorsport software stack~\cite{betz_tum_2022}.
The source code of all shown concepts is available at \url{https://github.com/TUMFTM/3DVehicleDynamicsStateEstimation}.

\begin{figure}[!t]
    \centering
    \vspace*{0.1cm}
    \includegraphics[width=\columnwidth]{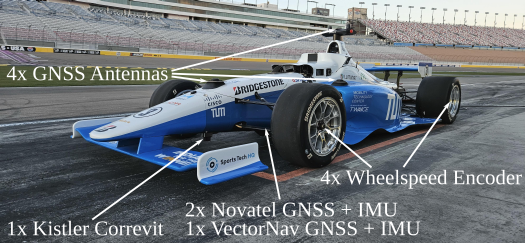}
    \caption{Sensor Setup of the Dallara AV-21 race car used for data recording on the Las Vegas Motorspeedway.}
    \label{fig: sensor-setup}
\end{figure}

Fig.~\ref{fig: sensor-setup} shows the Dallara AV-21 race car's sensor setup used to record the sensor data and evaluate the algorithms.
This comprises three GNSS receivers with four separate antennas, three independent Inertial Measurement Units (IMU), and four wheel-speed encoders. An additional Kistler SF-Motion Correvit Sensor delivers accurate longitudinal and lateral velocity measurements solely used for evaluation. Our main contributions are as follows:

\begin{itemize}
    \item We propose an integrated fusion scheme combining the stability of an Extended Kalman Filter (EKF) with a point-mass motion model and the accuracy of an Unscented Kalman Filter (UKF) with an integrated Single-Track Model (STM) to leverage model knowledge.
    \item We evaluate the online estimation of road angles with various strategies and show improvements in localization precision and lateral velocity estimation. We compare our results with a planar estimation scheme and an industry-grade INS solution.
    \item We show the impact of varying signal quality on the stability of the state estimate and suggest an improved fusion strategy to generate a smooth output.
\end{itemize}
\section{Related Work}
\label{sec:RelatedWork}

State estimation for autonomous vehicles inherits two main tasks.
First, the estimation of the pose (position and orientation), and second, the estimation of the vehicle's dynamic state (linear and angular velocities).
Most fusion strategies utilize variations of a Kalman Filter~\cite{kalman_new_1960} to leverage the advantages of different sensors.

Vehicle dynamics state estimation is essential, yet not limited, to autonomous vehicles. 
Therefore, various approaches exist to estimate the vehicle's longitudinal and lateral velocities.
These approaches can be divided by the motion model in point-mass~\cite{bechtloff_schatzung_2018, wischnewski_vehicle_2019, selmanaj_vehicle_2017} and vehicle-dynamics-model-based approaches~\cite{antonov_unscented_2011, wielitzka_state_2014}.
Some authors also use two separate estimators combining the advantages of point-mass and vehicle-dynamics-based methodologies~\cite{villano_cross-combined_2021}.
Furthermore, several data-driven algorithms have been developed~\cite{gonzalez_simultaneous_2020, ziaukas_estimation_2021}. 
Besides, combinations of model-based and data-driven methods exist~\cite{bertipaglia_unscented_2023}.
Chindamo et al., Jin et al., and Liu et al.~\cite{chindamo_vehicle_2018,jin_advanced_2019,liu_systematic_2023} provide a comprehensive overview of the different approaches.
Most authors neglect the impact of road slope and banking and are, thus, only applicable on approximately planar roads~\cite{bechtloff_schatzung_2018, chindamo_vehicle_2018}.

Bechtloff~\cite{bechtloff_schatzung_2018} uses an Unscented Kalman Filter (UKF) with a point-mass motion model.
The prediction is afterward corrected with the measured wheel speeds of the front axle. 
A virtual tire force measurement considering the yaw rate and the lateral acceleration stabilizes the estimation.
The calculated engine and brake torque are used to correct the estimation in the longitudinal direction.
The estimation also considers banked or inclined roads. 
Furthermore, he shows that a UKF outperforms an Extended Kalman Filter (EKF) for side slip angle estimation due to the high model nonlinearities.
Antonov et al.~\cite{antonov_unscented_2011} and Wielitzka et al.~\cite{wielitzka_state_2014} use a detailed vehicle dynamics motion model. 
Their comparison of EKF and UKF estimators yields the same conclusion as Bechtloff, showing the superior performance of the UKF compared to the EKF for side slip angle estimation.  
However, those approaches require a more thoughtful and time-consuming tuning than EKFs~\cite{chindamo_vehicle_2018, antonov_unscented_2011}. 
Yang et al.~\cite{yang_comparison_2017} compare EKF and UKF for different scenarios in autonomous vehicle navigation and show a superior performance of the UKF.
However, the UKF's stability cannot be guaranteed as a positive-definite covariance matrix cannot be ensured~\cite{yang_comparison_2017}. 

Besides the introduced estimation schemes, optical sensors can be used to measure the vehicle's lateral and longitudinal velocities accurately. 
Nevertheless, those are generally difficult to integrate into the vehicle packaging, costly, and not robust against environmental conditions.
Further, many competitions prohibit the use of such~\cite{chindamo_vehicle_2018}.
Data-driven methods have shown promising results in terms of side slip angle estimation. However, those require ground truth data for the initial training, might yield nonphysical behavior, and lack the ability to deal with alternated vehicle data~\cite{chindamo_vehicle_2018}.

Research considering a vehicle's pose estimation mainly focuses on using additional optical sensors, such as cameras and LiDARs.
Those are used to surpass shaded areas or completely substitute GNSS receivers.
However, RTK-GNSS, LiDARs, and visual odometry all lack a high-frequency update rate.
Consequently, several fusion strategies exist to incorporate various measurement sources.
Wan et al.~\cite{wan_robust_2018} utilize an Error State EKF (ES-EKF) to adaptively fuse IMU, GNSS, and LiDAR for shaded downtown areas.
Meguro et al.~\cite{meguro_low-cost_2018} demonstrate a pitch angle estimation considering the longitudinal velocity estimate and the measured acceleration.
Furthermore, they fuse the measured wheel speeds to increase estimation quality in urban scenarios.
Gao et al.~\cite{gao_improved_2022} show the beneficial impact of an additional velocity measurement considering the dead reckoning capabilities.
They switch between GNSS-based estimation and a vehicle dynamics model for shaded areas.
Robot Localization~\cite{menegatti_generalized_2016} provides open-source implementations of an EKF and a UKF for three-dimensional state estimation with point-mass motion models.
Autoware.Auto~\cite{Autoware} deliver an open-source EKF considering a two-dimensional state vector and a point-mass model. Furthermore, they implement an outlier rejection based on the distance of the measurement to the predicted state estimate.

In full-scale autonomous racing~\cite{betz_survey_2022}, Wischnewski et al.~\cite{wischnewski_vehicle_2019} suggest an EKF with a point-mass motion model and an additional Particle Filter to fuse LiDAR measurements.
They demonstrate superior performance compared to an STM. 
However, the estimation is limited to two-dimensional roads.
Lee et al.~\cite{lee_enhancing_2024} demonstrate resilient navigation fusing LiDAR and GNSS signals based on their applicability.
They directly switch between those measurements considering the current signal degradation.
Massa et al.~\cite{massa_lidar-based_2020} use a Particle Filter to estimate the vehicle's pose with LiDAR measurements and an EKF to deliver a smooth output signal.
Valls et al.~\cite{valls_design_2018} implement a LiDAR cone detection as a localization input. 
Afterward, they use an EKF to smooth the output.
Furthermore, they demonstrate the advantages of a signal delay compensation and a measurement outlier rejection.

\begin{figure*}[ht]
    \centering{\includegraphics[width=0.9\textwidth]{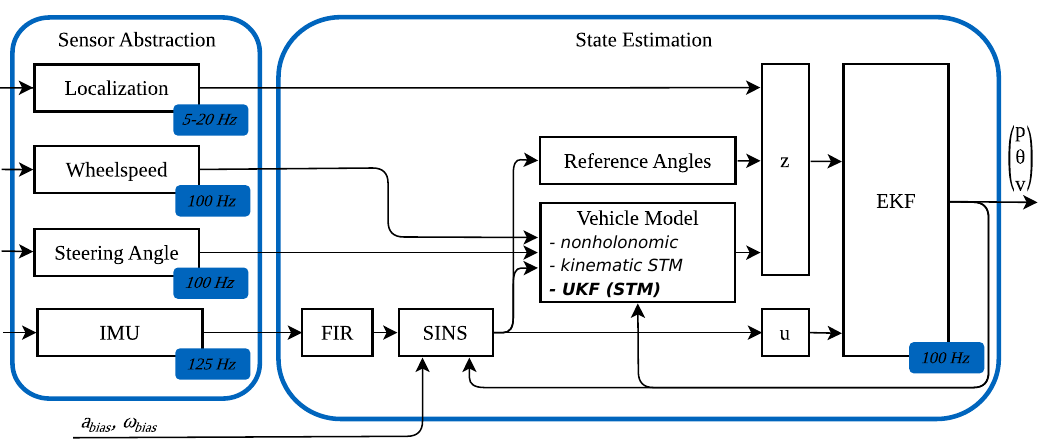}}
    \caption{High-level architecture of the state estimation software stack highlighting the core components and their interaction.}
    \label{fig:general_concept}
\end{figure*}

Adjusting the measurement or process noise covariance matrices can enhance the state estimate.
A modification based on the signal standard deviation significantly improves the root-mean-squared error (RMSE) of the position estimate~\cite{werries_adaptive_nodate}.
Similar results can be achieved using an Innovation Adaptive Estimation Adaptive Kalman Filter~\cite{liu_innovative_2018}.
Thereby, the variance of an innovation sequence is estimated and combined with an attenuation factor to focus on the most current measurements.
Yin et al.~\cite{yin_sensor_2023} use a Rauch-Tung-Striebel smoothing algorithm to enhance the robustness and smoothness of their predictions.

Few authors have shown the impact of road angles on the estimation accuracy. Yet, most approaches neglect road inclination and banking. Especially considering high velocities and accelerations, as observed in high-speed race cars, no approaches are known to the authors. The consequent nonlinearities are, therefore, not handled in current three-dimensional approaches.
Further, all localization approaches applied on real vehicles utilize an EKF due to the computational stability and neglect the performance gains achievable with a UKF.

Vehicle Motion Controllers react to deviations from the race line to reduce the cross-track error or - in case of Tube-based Controllers - to ensure staying within a defined driving tube~\cite{wischnewski2022}.
Consequently, positional jumps provoke unstable control behavior and have to be avoided.
Simply fusing the sensor signals after driving through areas without RTK-GNSS coverage or switching between different sources, as shown in the literature, has led to such positional jumps in the estimate in our experiments. No approaches are known to the authors to handle such positional jumps and enable a robust control behavior.
\section{Methodology}
\label{sec:Methodology}

\subsection{General Filter Design}

We follow the findings of~\cite{tseng_2007} by choosing an EKF  design over a UKF to avoid numerical instabilities. However, we utilize a UKF to estimate the vehicle side slip angle as an additional measurement input to leverage the superior performance. The modular concept ensures a robust state estimate at all times, while the UKF handles the higher nonlinearity of the vehicle model.
Our proposed fusion concept is depicted in Fig.~\ref{fig:general_concept} and is explained in detail subsequently.

First, we integrate an abstraction layer of the sensor interface, handling the signal validity and uniforming the sensor messages. This enables an easy deployment with different sensor setups without alternating the core state estimation.
The core state estimation module is based on a point-mass motion model to ensure robustness against falsified parametrization and applicability without accurate parameters.
The measurement vector $z$ contains positional $p_{meas}$ and orientation measurements $\theta_{meas}$ from a localization source. In our case, the RTK-GNSS receivers.
Further, we add reference angles and virtual velocity measurements $v_{meas}$ utilizing an STM as a vehicle model. 

\begin{equation}
    z=(p_{meas},\,\theta_{meas},\,v_{meas})^T \\
\end{equation}

The input vector $u$ consists of the averaged, Finite Impulse Response (FIR)-filtered IMU measurements of the angular velocities ${\omega_x}$, ${\omega_y}$, ${\omega_z}$ and the linear accelerations ${a_x}$, ${a_y}$, ${a_z}$. These are, furthermore, transformed to the center of gravity (COG) and corrected by sensor biases identified in stationary tests.

\begin{equation}
    u=(\omega_x,\,\omega_y,\,\omega_z,\,a_x,\,a_y,\,a_z)^T \\
\end{equation}

The state vector $x$ contains the vehicle's position $p$, orientation angles $\theta$, and linear velocities $v$. 

\begin{equation}
    x=(p,\,\theta,\,v)^T \\
\end{equation}

The state prediction of the EKF is based on the Euler Forward Integration of a constant velocity model. 
The derivate of the velocity ${\dot v}$ considering the angular velocities and the linear accelerations is shown in Eq.~\ref{eq_overall_accel_compensation}.
The constant gravity component is compensated by the vehicle's roll ${\phi}$ and pitch ${\theta}$ angle.

\begin{equation}
    \begin{aligned}
        \left[\begin{array}{c}{{\dot{v}_{x}}}\\ {{\dot{v}_{y}}}\\ {{\dot{v}_{z}}}\end{array}\right]
        =&-\left[\begin{array}{ccc}{{0}}&{{-\omega_{z}}}&{{\omega_{y}}}\\ {{\omega_{z}}}&{{0}}&{{-\omega_{x}}}\\{{-\omega_{y}}}&{{\omega_{x}}}&{{0}}\end{array}\right]\,\left[\begin{array}{c}{{{v}_{x}}}\\ {{{v}_{y}}}\\ {{{v}_{z}}}\end{array}\right]\\
        &+\left[\begin{array}{c}{{{a}_{x}}}\\ {{{a}_{y}}}\\ {{{a}_{z}}}\end{array}\right] + g\,\left[\begin{array}{c}{{\mathrm{sin}(\theta)}}\\ {{-\mathrm{sin}(\phi)\,\mathrm{cos}(\theta)}}\\ {{-\mathrm{cos}(\phi)\,\mathrm{cos}(\theta)}}\end{array}\right]
    \end{aligned}    
    \label{eq_overall_accel_compensation}
\end{equation}

The orientation angles are calculated using the following differential equation.
The race car inherits a stiff suspension system. Thus, the vehicle orientation is assumed to directly resemble the road inclination and banking. 

\begin{equation}
    \begin{aligned}
        \left[\begin{array}{c}{{\dot{\phi}}}\\ {{\dot{\theta}}}\\ {{\dot{\psi}}}\end{array}\right]
        =&\left[\begin{array}{ccc}{{1}}&{{\mathrm{sin}(\phi)\,\mathrm{tan}(\theta)}}&{{\mathrm{cos}(\phi)\,\mathrm{tan}(\theta)}}\\ {{0}}&{{\mathrm{cos}(\phi)}}&{{-\mathrm{sin}(\phi)}}\\{{0}}&{{\mathrm{sin}(\phi)\,\mathrm{sec}(\theta)}}&{{\mathrm{cos}(\phi)\,\mathrm{sec}(\theta)}} \end{array}\right]
        *\left[\begin{array}{c}{{\omega_{x}}}\\ {{\omega_{y}}}\\ {{\omega_{z}}}\end{array}\right]
    \end{aligned}
    \label{eq_attitude_kinematics}
\end{equation}

Those predictions are corrected using the sensor measurements to obtain the final state estimate. The predicted state and sensor measurement are weighed based on the assumed process and measurement covariances.

\subsection{Estimation of road geometries}

To stabilize the road angle estimation, we calculate a set of reference angles as suggested in \cite{tseng_2007}. We reformulate the longitudinal and lateral kinematics to stabilize the roll and pitch estimation, as shown in Eq.~\ref{eq_ref_phi} and \ref{eq_ref_theta}.

\begin{equation}
    {\phi_{ref} = asin\left(\frac{-\dot v_y + a_{y} - \omega_z\,v_x + \omega_x\,v_z}{g\,cos(\theta)}\right)}
    \label{eq_ref_phi}
\end{equation}
\begin{equation}
    {\theta_{ref} = asin\left(\frac{\dot v_x - a_{x} - \omega_z\,v_y + \omega_y\,v_z}{g}\right)}
    \label{eq_ref_theta}
\end{equation}

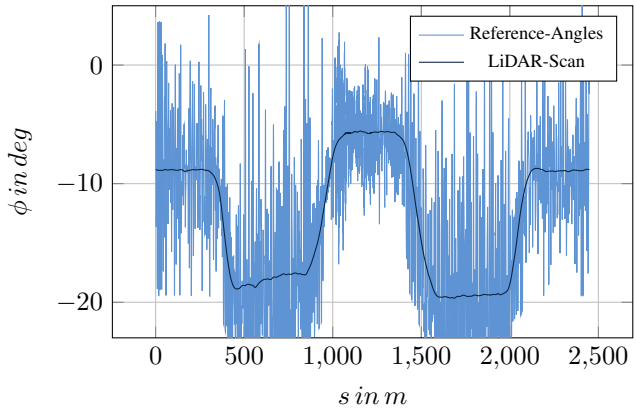
\begin{figure}[h]
	\definecolor{tum-blue-brand}{HTML}{3070B3}
\definecolor{tum-blue-dark}{HTML}{072140}
\definecolor{tum-orange}{HTML}{F7811E}
\definecolor{tum-grey-5}{HTML}{DDE2E6}
\definecolor{tum-blue-light}{HTML}{5E94D4}
\begin{tikzpicture}
    \begin{axis}[
        width=8.5cm,
        height=6cm,
        xlabel={$s\,in\,m$},
        ylabel={$\phi\,in\,deg$},
        ymin=-23, ymax=5,
        legend pos=north east,
        grid=both,
        legend style={font=\scriptsize},
    ]
    
    \addplot[tum-blue-light] table[x=/core/state/cpp/ekf_3d/s/data, y=ref_phi, col sep=comma, row sep=newline]{data/vegas_roll_pitch.csv};
    \addplot[tum-blue-dark] table[x=/core/state/cpp/ekf_3d/s/data, y=track_phi, col sep=comma, row sep=newline]{data/vegas_roll_pitch.csv};

    \legend{Reference-Angles, LiDAR-Scan}
    \end{axis}
\end{tikzpicture}
	\caption{Banking angle at LVMS based on aerial LiDAR measurements and calculated reference angles.}
	\label{fig: roll}
\end{figure}

Fig.~\ref{fig: roll} depicts the calculated reference angles on LVMS compared to aerial LiDAR scans.

\subsection{Virtual Velocity Measurement}

We integrate a virtual velocity measurement to improve the stability and accuracy of the estimation. 
Therefore, we compare three different approaches. 
First, we directly use the product of the wheel-speed measurements with the dynamic tire radius as a nonholonomic constraint. Next, we integrate the steering angle to calculate longitudinal and lateral velocity using a kinematic STM (kSTM) as shown in~\cite{gao_improved_2022, meguro_low-cost_2018}. Thereby, we calculate the kinematic side slip angle ${\beta_{kin}}$ considering the distance of the COG to the rear axle ${l_r}$ and the wheelbase ${l}$ as well as the steering angle ${\delta_f}$.

\begin{equation}
	\beta_{kin} = \mathrm{atan}\left(\dfrac{l_{r}}{l} \mathrm{tan(\delta_{f})} \right)
    \label{eq:bkin}
\end{equation}
\begin{equation}
    v_{x} = v \mathrm{cos}(\beta_{kin})
    \label{eq:vx}
\end{equation}
\begin{equation}
    v_{y} = v \mathrm{sin}(\beta_{kin})
    \label{eq:vy}
\end{equation}

Third, we integrate a UKF comprising an STM with a Pacejka Magic Formula tire model~\cite{pacejka2006} and, thus, the ability to handle nonlinear scenarios. 
The UKF is chosen based on the design presented in \cite{bechtloff_schatzung_2018}.

The road angle estimation is done in the EKF due to lower non-linearity than the vehicle side slip. Thus, we reduce the state vector compared to Bechtloff to the absolute velocity and the side slip angle. This also increases the flexibility of the entire system due to the easier tunability of EKFs compared to UKFs~\cite{chindamo_vehicle_2018, antonov_unscented_2011}.
The prediction update of the UKF is shown in Eq.~\ref{eq_vdot} and~\ref{eq_betadot}.

\begin{equation}
    \begin{aligned}
    \dot{v} &= \mathrm{cos}(\beta) (a_{x} + g \mathrm{sin}(\theta)) \\
    &\quad+ \mathrm{sin}(\beta) (a_{y} - g \mathrm{sin}(\phi) \mathrm{cos}(\theta))
    \label{eq_vdot}
    \end{aligned}
\end{equation}

\begin{equation}
    \begin{aligned}
    \dot{\beta} &= -\frac{\mathrm{sin}(\beta)(a_{x} + g \mathrm{sin}(\theta))}{v} \\
    &\quad+ \frac{\mathrm{cos}(\beta)(a_{y} - g \mathrm{sin}(\phi)\mathrm{cos}(\theta))}{v} - \omega_{z}
    \end{aligned}
    \label{eq_betadot}
\end{equation}

To correct the states with virtual measurements, the velocity at the middle point of the front axle ${v_{FA}}$ is calculated. Furthermore, the Pacejka Magic Formula tire model~\cite{pacejka2006} (Eq.~\ref{eq_MF}) is used to calculate the longitudinal and lateral tire forces of each axle. The respective longitudinal and lateral slip values $\sigma_x,\,\sigma_y$ at the front (FA) and rear axle (RA) are inputs to the tire model.

\begin{equation}
	\begin{array}{c@{\quad}c@{\quad}c}
		\boldsymbol{y} & = & \boldsymbol{h}(\boldsymbol{x}, \boldsymbol{u}) \\
		\begin{bmatrix} 
			v_{FA} \\
			v_{FA} \\
			F_{x, FA}\\
			F_{x, RA}\\
			F_{y, FA}\\
			F_{y, RA}\\		
		\end{bmatrix} & = & \begin{bmatrix} 
			v \mathrm{cos}(\delta_\mathrm{f} - \beta) + l_\mathrm{f} \omega_\mathrm{z} \mathrm{sin}(\delta_\mathrm{f}) \\
			v \mathrm{cos}(\delta_\mathrm{f} - \beta) + l_\mathrm{f} \omega_\mathrm{z} \mathrm{sin}(\delta_\mathrm{f}) \\
            F_\mathrm{z, FA} MF(\mathrm{\sigma_{x, FA}}) \\
            F_\mathrm{z, RA} MF(\mathrm{\sigma_{x, RA}}) \\
            F_\mathrm{z, FA} MF(\mathrm{\sigma_{y, FA}}) \\
            F_\mathrm{z, RA} MF(\mathrm{\sigma_{y, RA}}) \\
		\end{bmatrix}
	\end{array}
\end{equation}

\begin{equation}
    MF(x) = D \mathrm{sin}[C \mathrm{atan}(B \mathrm{x} - \\ E (B \mathrm{x} - \mathrm{atan}(B \mathrm{x})))]\\
    \label{eq_MF}
\end{equation}

The virtual measurements $z$ consist of calculating the vehicle speed at the middle of the front axle based on the wheel speeds of the front tires. Furthermore, the acting forces at each axle in the longitudinal direction are calculated based on the brake and engine torque. The lateral acceleration, the vehicle's inertial momentum, and the front axle's longitudinal force are employed to assess the lateral forces.
Interpolation between ${v_{FA}}$ for low excitation and the calculated tire forces for higher slip values is done by utilizing an adaptive covariance matrix~\cite{bechtloff_schatzung_2018}.

\begin{equation}
	\begin{array}{c@{\quad}c@{\quad}c}
		\boldsymbol{z} & = & \boldsymbol{g}(\boldsymbol{u}) \\
		\begin{bmatrix} 
			v_\mathrm{FA} \\
			v_\mathrm{FA} \\
			F_\mathrm{xT, f}\\
			F_\mathrm{xT, r}\\
			F_\mathrm{yT, f}\\
			F_\mathrm{yT, r}\\
		\end{bmatrix} & = & \begin{bmatrix} 
			\omega_\mathrm{fl} r_\mathrm{dyn, f} + \omega_\mathrm{z} \frac{b_\mathrm{f}}{2} \mathrm{cos}(\delta_\mathrm{f})   \\
			\omega_\mathrm{rl} r_\mathrm{dyn, f} - \omega_\mathrm{z} \frac{b_\mathrm{f}}{2} \mathrm{cos}(\delta_\mathrm{f})  \\
			(M_\mathrm{B, fl} + M_\mathrm{B, fr}) / r_\mathrm{dyn, f}\\
			(M_\mathrm{B, rl} + M_\mathrm{B, rr} + M_\mathrm{D}) / r_\mathrm{dyn, r}\\
			\dfrac{a_\mathrm{y} l_\mathrm{r}  m + \dot{\omega}_\mathrm{Z} J_\mathrm{z}}{l \mathrm{cos}(\delta_\mathrm{f})} - 
            \dfrac{\mathrm{sin}(\delta_\mathrm{f})}{\mathrm{cos}(\delta_\mathrm{f})}
             F_\mathrm{xT, f}\\
			\dfrac{a_\mathrm{y} l_\mathrm{f}  m + \dot{\omega}_\mathrm{Z} J_\mathrm{Z}}{l}\\	
		\end{bmatrix}
	\end{array}
\end{equation}

\subsection{Alternating Signal Quality}

\begin{figure}[h]
    \centerline{\includegraphics[width=0.8\linewidth]{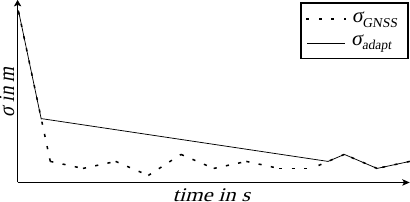}}
	\caption{Covariance adaptation strategy after a signal dropout. ${\sigma_{GNSS}}$ symbolizes the reported standard deviation of the GNSS receiver, while ${\sigma_{adapt}}$ represents the standard deviation used for the adaptive covariance of the EKF.}
	\label{fig:monza_covadapt}
\end{figure}

As mentioned, positional jumps in the estimation can lead to unstable control behavior.
We developed two strategies to avoid those in case of alternating signal quality. 
First, we fuse measurements based on their reported standard deviation ${\sigma_{GNSS}}$ and past measurement history. 
This allows us to weigh how much each measurement influences the system state. 
Once ${\sigma_{GNSS}}$ decreases after a previous signal drop and reaches a specified threshold, the covariance of the measurement is not equated to ${\sigma_{GNSS}}$ anymore. Instead, we linearly decay the covariance until ${\sigma_{GNSS}}$ is reached and afterward follow ${\sigma_{GNSS}}$. This ensures a smooth transition towards the measurement instead of positional jumps. The linear decay has shown the best results within the tried functions. Fig.~\ref{fig:monza_covadapt} depicts the implemented covariance adaption strategy.

Second, we employ an outlier rejection based on the Mahalonobis distance~\cite{mahalanobis1936}.
Thereby, the distance of the measured position to the prediction of the EKF is calculated.
The measurement value is then clamped to a maximum value if the distance exceeds a defined limit.
The Mahalonobis distance is calculated in the vehicle frame, enabling orthogonal and longitudinal boundaries to the current race line.
Consequently, boundaries towards the track bounds can be tuned more carefully.
Subsequently, the combination of adaptive covariance and Mahalonobis distance outlier rejection is called ACOR.
\section{Results}
\label{sec:Results}

\subsection{Experimental Setup}

We evaluate our newly designed state estimation on a real autonomous race car in two different race tracks, LVMS and MON.
The used Dallara AV-21 race car, shown in Fig.~\ref{fig: sensor-setup}, contains three RTK-GNSS receivers, three IMUs, and four wheelspeed encoders. Furthermore, the vehicle model uses a steering angle sensor, brake pressure, and engine torque.
The RTK-GNSS receivers provide an accurate localization measurement at frequencies of \SI{20}{\hertz} and \SI{5}{\hertz}. 
However, multi-path interference, signal blockage, and large distances to the reference station can strongly degrade the measurement quality~\cite{groves_knovel_2013}. 
The IMU sensors, with a frequency of \SI{125}{\hertz}, do not suffer from those environmental effects. 
Their position estimate is limited to short-term periods due to the integration of process drift caused by sensor misplacement, biases, or random signal drifts~\cite{groves_knovel_2013, quinchia_comparison_2013}. 
Lastly, the wheel-speed encoders update at \SI{100}{\hertz} and offer an additional measurement source of the vehicle's speed.
Their primary error results from varying slip conditions and dynamical tire diameters.

To provide a fair comparison with a state-of-the-art estimation scheme and an industry-grade INS solution, we only fuse two of the three GNSS receivers and use the third one solely for evaluation purposes. 
Further, we use an optical velocity sensor to assess the velocity estimation capabilities. Aerial LiDAR scans serve as a ground truth for the road angle estimation. Thereby, we linearize the banking angle orthogonal to the track by connecting both track bounds and calculate the present road angle based on the vehicle's current yaw angle.
Due to the proven performance in autonomous racing, and thus at high velocities and accelerations, we choose the planar state estimation suggested by Wischnewski et al.~\cite{wischnewski_vehicle_2019} as a comparison in our studies. This approach is subsequently named 2D-EKF.
All filters have been tuned individually to achieve their best performance.

The real-time capability of all algorithms has been successfully tested on a dSpace AUTERA AutoBox, installed in the Dallara AV-21 racecar, with an Intel Xeon CPU (12 x 2.0 GHz) and 32 GB RAM. The 2D-EKF has an average run-time of 66.8 microseconds, the 3D-EKF has an average run-time of 160.8 microseconds and the UKF-based side slip angle estimator an average run-time of 55.7 microseconds.

\subsection{Road and Side Slip Angle Estimation}

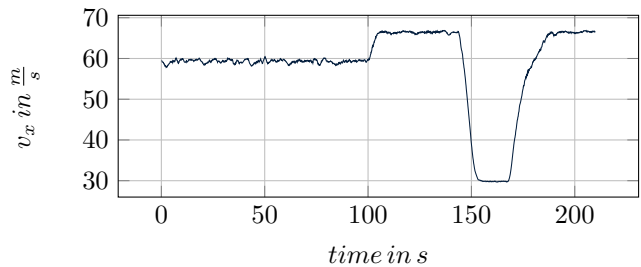
\begin{figure}[h]
	\definecolor{tum-blue-brand}{HTML}{3070B3}
\definecolor{tum-blue-dark}{HTML}{072140}
\definecolor{tum-orange}{HTML}{F7811E}
\definecolor{tum-grey-5}{HTML}{DDE2E6}
\definecolor{tum-blue-light}{HTML}{5E94D4}

\begin{tikzpicture}
    \begin{axis}[
        width=8.5cm,
        height=4cm,
        xlabel={$time\,in\,s$},
        ylabel={$v_x\,in\,\frac{m}{s}$},
        legend pos=south west,
        grid=both,
        legend style={font=\scriptsize},
    ]

    \addplot[tum-blue-dark] table[x=timestamp_s, y=/kistler/correvit/vel_x, col sep=comma, row sep=newline]{data/vegas24_kistler_cut.csv};

    \end{axis}
\end{tikzpicture}
	\caption{Longitudinal velocity profile at LVMS.}
	\label{fig:vegas_long}
\end{figure}

Side slip angle estimation is especially challenging considering high banking angles~\cite{bechtloff_schatzung_2018, chindamo_vehicle_2018}. Therefore, we compare the velocity estimation of all concepts at LVMS, where banking angles of up to \SI{20}{\degree} are reached. Further, we include one braking and one acceleration zone to consider different slip states, as shown in Fig.~\ref{fig:vegas_long}.

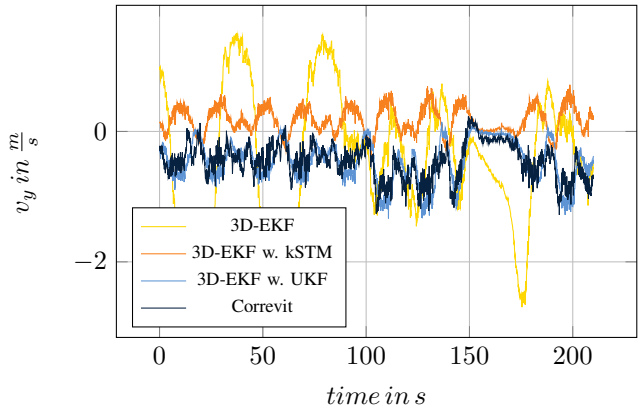
\begin{figure}[h]
	\definecolor{tum-blue-brand}{HTML}{3070B3}
\definecolor{tum-blue-dark}{HTML}{072140}
\definecolor{tum-orange}{HTML}{F7811E}
\definecolor{tum-grey-5}{HTML}{DDE2E6}
\definecolor{tum-blue-light}{HTML}{5E94D4}
\definecolor{tum-yellow}{RGB}{254, 215, 2}

\begin{tikzpicture}
    \begin{axis}[
        width=8.5cm,
        height=6cm,
        xlabel={$time\,in\,s$},
        ylabel={$v_y\,in\,\frac{m}{s}$},
        legend pos=south west,
        grid=both,
        legend style={font=\scriptsize},
    ]

    \addplot[tum-yellow] table[x=timestamp_s, y=/core/state/cpp/ekf_3d/odometry/twist/twist/linear/y, col sep=comma, row sep=newline]{data/vegas24_kin_cut.csv};
    \addplot[tum-orange] table[x=timestamp_s, y=/core/state/cpp/ekf_3d/odometry/twist/twist/linear/y, col sep=comma, row sep=newline]{data/new_vegas24_semi_kstm_cut.csv};
    \addplot[tum-blue-light] table[x=timestamp_s, y=/core/state/cpp/ekf_3d/odometry/twist/twist/linear/y, col sep=comma, row sep=newline]{data/vegas24_ukf_cut.csv};
    \addplot[tum-blue-dark] table[x=timestamp_s, y=/kistler/correvit/vel_y, col sep=comma, row sep=newline]{data/vegas24_kistler_cut.csv};

    \legend{3D-EKF, 3D-EKF w. kSTM, 3D-EKF w. UKF, Correvit}
    \end{axis}
\end{tikzpicture}
	\caption{Lateral velocity estimates at LVMS of the baseline 3D-EKF and extensions with a kSTM and a UKF. An SF-Motion Correvit sensor measures the Ground Truth.}
	\label{fig: vegas_lat}
\end{figure}

The estimated lateral velocities of the baseline 3D-EKF and the extensions with a kSTM and a UKF are compared to the measurements of an optical velocity sensor in Fig.~\ref{fig: vegas_lat}. As can be seen, the baseline and the kSTM suggested in ~\cite{gao_improved_2022, meguro_low-cost_2018} do not yield accurate estimates. The UKF extension outperforms both approaches.

Tab.~\ref{tab:eval_speed} compares the previously introduced algorithm and its extensions, considering their capabilities for longitudinal and lateral speed estimation and road angle estimation at LVMS.
The comparison with the 2D-EKF is enabled by using the banking angle from a 3D map to correct the lateral acceleration measurements by the gravitational force.
Considering lateral velocity and side slip angle estimation, the 2D-EKF with an underlying banking map and the INS solution outperform the baseline 3D-EKF.
However, using a road angle map in the 3D-EKF as an additional measurement shows superior performance compared to 2D-EKF and INS.
Introducing reference angles significantly improves the lateral velocity and the road angle estimation compared to the baseline.
Including a virtual velocity measurement combined with reference angles further increases the estimation performance of lateral velocity and road angles. 
The best results are achieved utilizing reference angles and the previously introduced UKF-based side slip angle estimation.

\begin{table*}[ht]
    \caption{LVMS - Comparison of the various filter extensions considering velocity and road angle estimation errors.}
    \begin{center}
    \bgroup
    \def\arraystretch{1.2}
    \begin{tabular}{|c|l|l|c|c|c|c|c|c|c|c|c|c|}
        \hline
    \multicolumn{3}{|c|}{\multirow{2}{*}{\textbf{Method}}}& \multicolumn{2}{|c|}{$\mathbf{v_x\,in\,m/s}$}& \multicolumn{2}{|c|}{$\mathbf{v_y\,in\,m/s}$}& \multicolumn{2}{|c|}{$\pmb{\beta\,in\,\deg}$}& \multicolumn{2}{|c|}{$\pmb{\Theta\,in\,\deg}$}& \multicolumn{2}{|c|}{$\pmb{\Phi\,in\,\deg}$} \\
    \cline{4-13}
    \multicolumn{3}{|c|}{}& \textbf{RMSE}& \textbf{MAX}& \textbf{RMSE}& \textbf{MAX}& \textbf{RMSE}& \textbf{MAX}& \textbf{RMSE}& \textbf{MAX}& \textbf{RMSE}& \textbf{MAX} \\ \hline
    \multicolumn{3}{|c|}{INS} & 0.41 & 1.65 & 0.85 & 1.50 & 0.80 & 1.33 & 0.38 & \textbf{0.94} & 0.80 & 2.72 \\ \hline
    \multicolumn{3}{|c|}{2D-EKF, banking} & 0.38 & 1.41 & 1.12 & 2.05 & 1.08 & 1.86 & - & - & - & - \\ \hline
    \multirow{6}{*}{\rotatebox[origin=c]{90}{3D-EKF}}
    & \multicolumn{2}{|l|}{Vanilla} & 0.36 & \textbf{1.40} & 1.27 & 2.75 & 1.21 & 2.63 & 0.99 & 2.95 & 1.41 & 3.87 \\
    & \multicolumn{2}{|l|}{Road Angles} & 0.38 & 1.55 & 0.48 & 1.53 & 0.46 & 1.32 & - & - & - & - \\
    \cline{2-13}
    & \multirow{4}{*}{\rotatebox[origin=c]{90}{Ref. Angles}}
    & Vanilla & 0.37 & 1.53 & 0.80 & 2.15 & 0.81 & 2.07 & 0.64 & 1.56 & 1.17 & 3.10 \\
    & \multirow{4}{*}{} & Nonh. & 0.31 & 1.44 & 0.37 & 1.61 & 0.36 & 1.55 & 0.33 & 1.18 & 0.88 & 2.92 \\
    & \multirow{4}{*}{} & kSTM & 0.32 & 1.44 & 0.73 & 1.65 & 0.67 & 1.42 & 0.36 & 1.24 & 0.96 & 3.15 \\
    & \multirow{4}{*}{} & \textbf{UKF (Ours)} & \textbf{0.30} & 1.42 & \textbf{0.22} & \textbf{0.64} & \textbf{0.21} & \textbf{0.60} & \textbf{0.32} & 
    0.96 & \textbf{0.76} & \textbf{2.38} \\ \hline
    \end{tabular}
    \label{tab:eval_speed}
    \egroup        
\end{center}
\end{table*}

\subsection{Varying Signal Quality}

\begin{figure}[h]
    \centerline{\includegraphics[width=\linewidth]{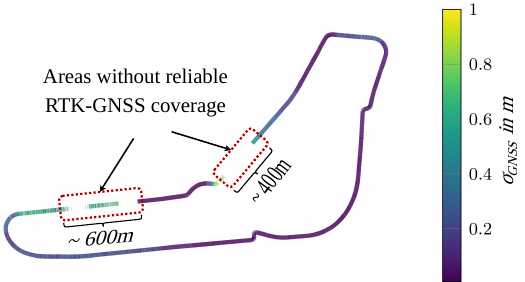}}
	\caption{GNSS standard deviation averaged over all GNSS sensors for the final race at MON in June 2023 (areas with a standard deviation larger than \SI{1}{\meter} are excluded).}
	\label{fig:monza_stddev}
\end{figure}

MON shows low banking angles with maximum values of \SI{4}{\degree}.
However, the GNSS signal shows high standard deviations caused by surrounding trees, underpasses, and metallic posts passing the track. 
Fig.~\ref{fig:monza_stddev} shows the averaged standard deviation over all three GNSS sensors ${\sigma_{GNSS}}$.
We excluded areas with a standard deviation higher than \SI{1}{\meter}.

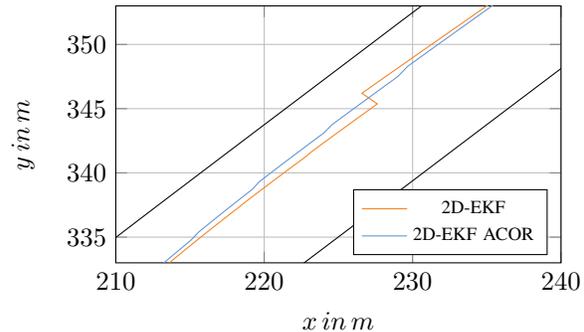
\begin{figure}[h]
	\definecolor{tum-blue-dark}{HTML}{072140}
\definecolor{tum-orange}{HTML}{F7811E}
\definecolor{tum-blue-brand}{HTML}{3070B3}
\definecolor{tum-blue-light}{HTML}{5E94D4}

\begin{tikzpicture}
    \begin{axis}[
        width=7.5cm,
        height=5cm,
        xlabel={$x\,in\,m$},
        ylabel={$y\,in\,m$},
        legend pos=south east,
        grid=both,
        legend style={font=\scriptsize},
        xmin=210,
        xmax=240,
        ymin=333,
        ymax=353,
    ]
    
    \addplot[black, forget plot] table[x=x_left_bound, y=y_left_bound, col sep=comma, row sep=newline]{data/monza_track.csv};
    \addplot[black, forget plot] table[x=x_right_bound, y=y_right_bound, col sep=comma, row sep=newline]{data/monza_track.csv};

    \addplot[tum-orange] table[x=/core/state/cpp/ekf_2d/odometry/pose/pose/position/x, y=/core/state/cpp/ekf_2d/odometry/pose/pose/position/y, col sep=comma, row sep=newline]{data/monza_acor.csv};
    \addplot[tum-blue-light] table[x=acor/core/state/cpp/ekf_2d/odometry/pose/pose/position/x, y=acor/core/state/cpp/ekf_2d/odometry/pose/pose/position/y, col sep=comma, row sep=newline]{data/monza_acor.csv};
    \addlegendentry{2D-EKF}
    \addlegendentry{2D-EKF ACOR}
    \addlegendentry{gt}
    \end{axis}
\end{tikzpicture}
	\caption{Positional jump with and without covariance adaptation and outlier rejection (ACOR).}
	\label{fig:monza_posjump}
\end{figure}

Directly switching back to the RTK-GNSS signal once the standard deviation reaches an adequate threshold yields positional jumps of the state estimation.
As mentioned, those provoke unstable control behavior and must be avoided.
Fig.~\ref{fig:monza_posjump} depicts such a positional jump shortly after the return of a GNSS measurement after a previous signal drop. The introduced application of ACOR yields smooth convergence of the estimate and has enabled stable control behavior during the testing sessions at MON.

\subsection{Localization Accuracy}

We evaluate the accuracy of the position based on a separate RTK-GNSS receiver that is not utilized in the Kalman Filters.
Since the maximum positional error is more critical orthogonal to the vehicle frame and, thus, toward the track bounds, we focus on the lateral positional error expressed in Frenet Coordinates.
As shown in Tab.~\ref{tab:eval_pos}, the 2D-EKF without ACOR shows the lowest deviations. Nevertheless, the high lateral position jumps have lead to unstable controller behavior in our experiments.
The proposed filter containing reference angles, ACOR, and a UKF outperforms all other solutions, including the INS solution and the 2D-EKF with ACOR. 
At LVMS, all shown filters show good results. Only the INS solution shows high lateral deviations up to \SI{0.94}{\meter}.

\begin{table}[ht]
    \caption{MON, LVMS - Positional Error orthogonal to the vehicle frame ($d_{err}\, in\,m$).}
    \begin{center}
    \bgroup
    \def\arraystretch{1.2}
    \begin{tabular}{|c|l|l|c|c|c|c|}
        \hline
        \multicolumn{3}{|c|}{\multirow{2}{*}{\textbf{Method}}}&\multicolumn{2}{|c|}{\textbf{MON}}&\multicolumn{2}{|c|}{\textbf{LVMS}} \\
        \cline{4-7}
        \multicolumn{3}{|c|}{}&\multicolumn{1}{|c|}{\textbf{RMSE}}&\multicolumn{1}{|c|}{\textbf{MAX}}&\multicolumn{1}{|c|}{\textbf{RMSE}}&\multicolumn{1}{|c|}{\textbf{MAX}} \\ \hline        
        \multicolumn{3}{|c|}{INS} &         0.27 & 0.67 & 0.25 & 0.94 \\ \hline
        \multirow{3}{*}{\rotatebox[origin=c]{90}{2D-EKF}} 
        & \multicolumn{2}{|l|}{Vanilla} & \textbf{0.07} & \textbf{0.20} & - & - \\ 
        & \multicolumn{2}{|l|}{ACOR} & 0.10 & 0.36 & - & - \\
        & \multicolumn{2}{|l|}{ACOR, banking} & 0.08 & 0.29 & 0.06 & 0.23 \\ \hline
        \multirow{6}{*}{\rotatebox[origin=c]{90}{3D-EKF (ACOR)}}
        & \multicolumn{2}{|l|}{Vanilla} & 0.10 & 0.34 & 0.09 & 0.21 \\
        & \multicolumn{2}{|l|}{Road Angles} & 0.08 & 0.28 & 0.07 & 0.18 \\
        \cline{2-7}
        & \multirow{4}{*}{\rotatebox[origin=c]{90}{Ref. Angles}}
        & Vanilla & 0.10 & 0.52 & 0.08 & 0.19 \\
        & \multirow{4}{*}{} & Nonh. & 0.08 & 0.43 & \textbf{0.04} & 0.12 \\
        & \multirow{4}{*}{} & kSTM & 0.08 & 0.39 & \textbf{0.04} & \textbf{0.11} \\
        & \multirow{4}{*}{} & \textbf{UKF (Ours)} & 0.08 & 0.28 & \textbf{0.04} & \textbf{0.11} \\ \hline
    \end{tabular}
    \egroup
    \label{tab:eval_pos}
\end{center}
\end{table}

\subsection{Limitations}

The introduced state estimation has been evaluated on a Dallara AV-21 race car with a stiff suspension. 
Other vehicles may show higher roll and pitch movements of the chassis, which must be compensated to estimate the road geometry. 
Furthermore, the tire parameters have been adjusted to match the environmental conditions. 
Alternating weather conditions and tire temperatures and, thus, varying tire parameters have not been considered within our evaluation.
Another limitation is the tuning dependency of each filter. Even though over 1300 simulations have been performed to fine-tune the algorithms in several optimization loops, a global optimum cannot be guaranteed.
The banking angle evaluation neglects progressive banking at LVMS. Thus, the quantitative results could slightly deviate from reality.
We solely evaluate the localization performance in the case of present RTK corrections of the third receiver to decrease the impact of measurement noise on the ground truth. Consequently, the maximum lateral deviation might be higher in areas without GNSS coverage.
\section{Conclusion}
\label{sec:conclusion}

In this paper, we suggested a three-dimensional state estimation with the extension of reference angles and several virtual velocity measurements. We demonstrated the superior performance of our introduced state estimation compared to a state-of-the-art two-dimensional state estimation and an industry-grade INS solution. The evaluation was done on two racetracks with high velocities, accelerations, and different characteristics. LVMS shows high banking angles of up to \SI{20}{\degree}, while MON challenges the estimation due to multiple shaded areas and, thus, degraded GNSS signal quality. Our presented state estimation can estimate the road angles online and accurately predict the vehicle's side slip angle. A combination of reference angles and an additional UKF side slip angle estimator outperformed all other approaches in our comparisons.
Furthermore, we highlighted the importance of a smooth positional estimate for specific control algorithms. We developed a strategy to fuse the GNSS signal adaptively based on an adaptive covariance in combination with a Mahalonobis distance outlier rejection and demonstrated the resulting smooth convergence of the state estimation.
The limitations of this study have been shown and are currently being investigated. Our next steps are the integration of a parameter estimation in the UKF as suggested in \cite{wielitzka_state_2014,bechtloff_schatzung_2018} to improve the prediction for alternating grip values caused by different tire temperatures or track conditions. We are currently integrating a LiDAR localization measurement to decrease the GNSS dependency further.
The modular design of the developed algorithm enables transferability to different vehicles. However, further studies are needed to investigate the performance on passenger vehicles.

\section*{ACKNOWLEDGMENT}

We thank the IAC and the TUM Autonomous Motorsports team for their support during the data acquisition. We thank Kistler for lending us the SF-Motion Correvit sensor to validate our results. Moreover, thanks to Leica Geosystems and VectorNav Technologies for their support.

As the first author, Sven Goblirsch initiated and designed the paper's structure. He contributed to designing and tuning the proposed filters and implemented the side slip angle estimator.
Marcel Weinmann contributed to designing and implementing the demonstrated EKF and its extensions as part of his guided research project.
Johannes Betz made an essential contribution to the concept of the research project. He revised the paper critically for important intellectual content. Johannes Betz gives final approval for the version to be published and agrees to all aspects of the work. As a guarantor, he accepts responsibility for the overall integrity of the paper.

\bibliographystyle{IEEEtran}
\bibliography{IEEEabrv, main.bbl}

\end{document}